\documentclass[letterpaper, 10 pt, conference]{ieeeconf}  

\IEEEoverridecommandlockouts                              

\usepackage{graphics} 
\usepackage{epsfig} 
\usepackage{mathptmx} 
\usepackage{times} 
\usepackage{amsmath} 
\usepackage{amssymb}  
\usepackage{color,soul}

\title{\LARGE \bf
Deep Learning-Based Connector Detection for Robotized Assembly of Automotive Wire Harnesses*
}

\author{
Hao Wang$^{1}$ and Björn Johansson$^{1}$
\thanks{
*This work was supported by the Swedish innovation agency, Vinnova, and the strategic innovation program, Produktion2030, under grant no. 2022-01279.
The work was carried out within Chalmers’ Area of Advance Production.
The connectors used in this study were provided by Wiretronic AB.
The support is gratefully acknowledged.
}
\thanks{$^{1}$Hao Wang and Björn Johansson are with the Division of Production Systems, Department of Industrial and Materials Science, Chalmers University of Technology, Hörsalsvägen 7A, SE-412 96 Gothenburg, Sweden {\tt\small haowang@chalmers.se}}%
}

\begin{document}

\maketitle
\thispagestyle{empty}
\pagestyle{empty}

\begin{abstract}

The shift towards electrification and autonomous driving in the automotive industry makes automotive wire harnesses increasingly more critical for various functions of automobiles, such as maneuvering, driving assistance, and safety system.
It leads to more and more wire harnesses installed in modern automobiles, which stresses the great significance of guaranteeing the quality of automotive wire harness assembly.
The mating of connectors is essential in the final assembly of automotive wire harnesses due to the importance of connectors on wire harness connection and signal transmission.
However, the current manual operation of mating connectors leads to severe problems regarding assembly quality and ergonomics, where the robotized assembly has been considered, and different vision-based solutions have been proposed to facilitate the robot control system's better recognition of connectors.
Nonetheless, there has been a lack of deep learning-based solutions for detecting wire harness connectors in previous studies.
This paper presents a deep learning-based connector detection for robotized automotive wire harness assembly.
A dataset of twenty types of automotive wire harness connectors was created to train and evaluate a two-stage object detection model and a one-stage object detection model, respectively.
The experiment results indicate the effectiveness of deep learning-based connector detection for automotive wire harness assembly but are limited by the design of the exteriors of connectors.

\end{abstract}

\section{INTRODUCTION}
\label{sec:intro}

Electrification and autonomous driving have driven a paradigm shift in the current automotive industry, making the electronic system increasingly critical in modern automobiles.
Numerous automotive wire harnesses have been installed in current vehicles as an essential infrastructure for supporting signal transmission within the electronic system.
Meanwhile, more and more wire harnesses are expected to be installed, considering the increase of automotive wire harnesses in vehicles in the past decades and the paradigm shift in the industry.
Thus, it is crucial to guarantee the quality of the assembly of automotive wire harnesses.

\begin{figure}[t]
    \centering
    \includegraphics[width=\linewidth]{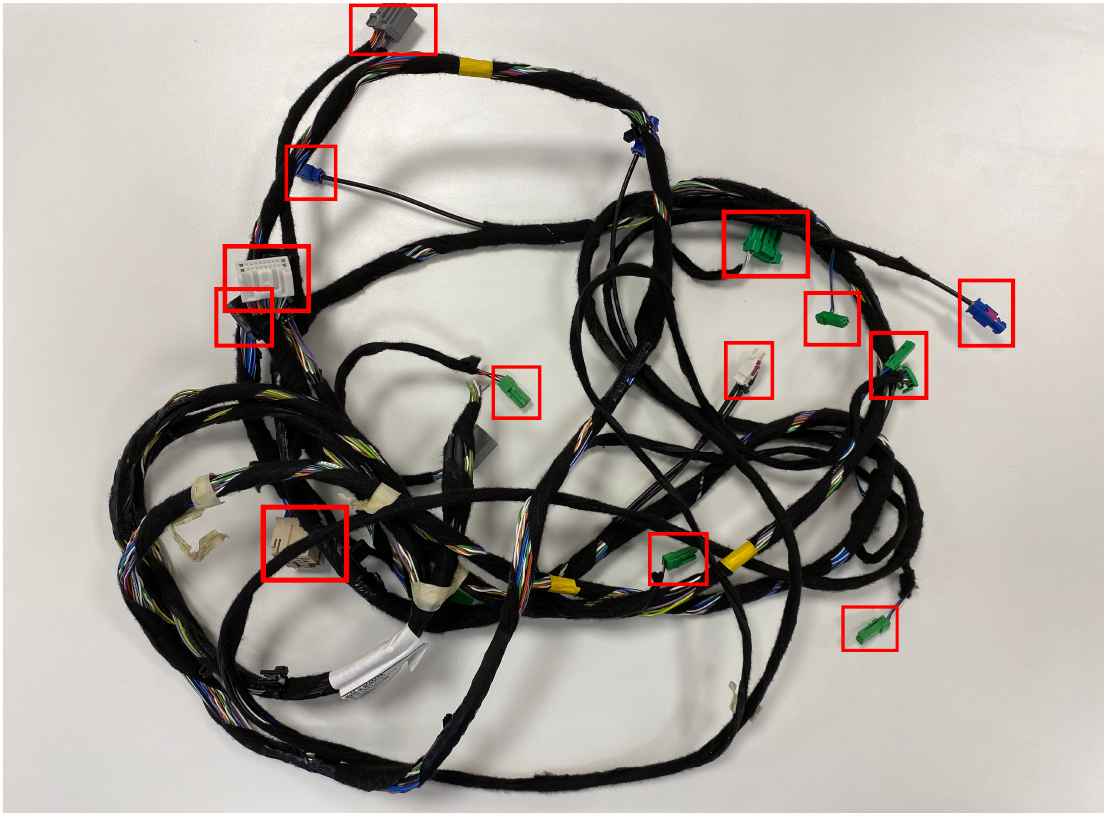}
    \caption{An example of an automotive wire harness, with connectors highlighted by red rectangles.}
    \label{fig:wire_example}
\end{figure}

However, the current final assembly of automotive wire harnesses into vehicles remains mostly manual and skill-demanding, which makes it challenging to control and improve the quality and productivity of the assembly.
Some manual operations also involve heavy lifting (for example, approximately 40 kg for some low-voltage automotive wire harnesses) and high-pressure manual manipulations on different components of automotive wire harnesses, which poses severe ergonomic problems to human operators.
In particular, the mating of connectors is one of the sub-process relating to ergonomic issues due to the repetitive high-pressure manual pressing in the assembly line.
Fig.~\ref{fig:wire_example} demonstrates an example of an automotive wire harness, where red rectangles highlight connectors on the automotive wire harness.

Connectors are essential components on automotive wire harnesses, among the others, such as clamps and cables.
Automotive wire harnesses are connected to the target unit or other automotive wire harnesses via connectors so the signal can be transmitted continuously within the electronic systems responsible for various functions of automobiles, which are safety-critical in particular.
Thus, ensuring the quality of mating connectors in the final assembly of wire harnesses into vehicles is critical.
However, the current manual process of mating connectors constrains the productivity and quality of assembly and generates ergonomic problems for human operators.
To relieve the problems regarding productivity, assembly quality, and ergonomics, robotized wire harness assembly is of great interest to the automotive industry, considering its better replicability, transparency, and comprehensibility, and has been discussed in different studies previously~\cite{koo2008development,koo2011development,jiang2011robotized,jiang2015robotized}.
Nevertheless, the robotized mating of connectors is non-trivial as the robotic operation needs to address not only high manipulation accuracy but also intricate structures and non-rigid materials of connectors~\cite{sun2010robotic}.
It is also fundamental to retrieve the geometrical information of connectors beforehand so that a robot arm can flexibly reach, grasp, and manipulate the perceived connector.

Computer vision has demonstrated a significant potential on the robotized assembly to the manufacturing industry in solving ergonomic issues while increasing quality and productivity~\cite{zhou2022computer}.
Previously, there have also been studies on computer vision techniques for robotized manipulation of wire harness connectors~\cite{sun2010robotic,di2009hybrid,di2012vision,tamada2013high,song2017electric,yumbla2020preliminary,zhou2020practical}.
However, a few studies discussed the task of connector detection~\cite{tamada2013high,yumbla2020preliminary,zhou2020practical}, where methods based on basic image processing techniques were mainly explored~\cite{tamada2013high,yumbla2020preliminary}.
Considering the various designs of connectors on automotive wire harnesses, such as colors, shapes, and sizes, it is intricate to manage the manual feature engineering on connectors for flexible robotized manipulation.
The recent advancement in implementing convolutional neural networks (CNN) and deep learning in computer vision research has demonstrated the extraordinary effectiveness of learning-based solutions for object detection compared to traditional image processing-based solutions~\cite{lecun2015deep}.
Zhou et al.~\cite{zhou2020practical} have previously explored deep learning-based connector detection for the robotized wire harness connection, but the proposal mainly focused on one-connector detection.
The learning-based detection on multiple connectors remained unsolved but is required for the robotized assembly of automotive wire harnesses in actual production.

This paper presents a study on the deep learning-based connector detection for the robotized mating of connectors on automotive wire harnesses and discusses the feasibility and potential problems of implementing deep learning-based object detection on the task of mating connectors in robotized automotive wire harness assembly under laboratory conditions.
As there is no publicly available dataset on automotive wire harness connectors, a dataset comprising twenty different types of connectors was collected initially.
Then, two different detection models, a two-stage object detection model, Faster R-CNN~\cite{ren2015faster}, and a one-stage object detection model, YOLOv5~\cite{yolov5}, were adopted for the training and inference.
The experiment results demonstrate the effectiveness of deep learning-based connector detection as both detection methods achieved remarkable detection outcomes with various combinations of connectors presenting in the scene.
Yet, detection performance can be improved further, and a more extensive dataset comprising more connectors and more images per connector is needed.
Some detection errors on classes and positions of connectors in inference results further reflect the effect of the design of the exterior of connectors, which motivates the future connector detection based on multi-view images of connectors and with new exterior design of connectors so that more visually distinguishable features of the connector can be extracted.

This paper is organized in the following structure:
Section~\ref{sec:related} introduces the related research in connector detection and deep learning-based object detection.
Section~\ref{sec:dataset} introduces the data collection and annotation strategy and the statistics of the collected dataset of connectors.
Section~\ref{sec:exp} introduces the experiment setups of two-stage and one-stage connector detection, whose results are presented and further discussed in Section~\ref{sec_result}.
The study is concluded in Section~\ref{sec:conclu} with an outlook on the future work of this study.

\section{RELATED WORK}
\label{sec:related}

\subsection{Connector Detection for Robotized Mating of Connectors}

Connector detection is needed to acquire the position and categories of connectors so that the robot can flexibly reach, grasp, and manipulate connectors.
Although some vision-based solutions have been proposed for facilitating different sub-tasks in robotized mating of connectors~\cite{sun2010robotic,di2009hybrid,di2012vision,tamada2013high,song2017electric,yumbla2020preliminary,zhou2020practical}, connector detection has yet gathered few attention in previous studies~\cite{tamada2013high,yumbla2020preliminary,zhou2020practical}, where the basic image processing-based methods are dominant~\cite{tamada2013high,yumbla2020preliminary}.

Tamada et al.~\cite{tamada2013high} proposed to recognize the types and poses of connectors using a high-speed vision system.
An image processing method was adopted in Tamada et al.~\cite{tamada2013high} to detect the positions of connectors via detecting the corners of connectors, which was further processed to calculate the orientations of connectors.
Yumbla et al.~\cite{yumbla2020preliminary} later proposed a basic image processing-based method to detect multiple connectors, including converting color space and applying color thresholding.
However, the task in Yumbla et al.~\cite{yumbla2020preliminary} was a one-class detection, where all connectors were considered the same class.
Deep learning-based connector detection has also been discussed in a recent study~\cite{zhou2020practical}, which proposed to roughly locate the position of a connector and then zoom in to the detected connector to acquire the finer pose of the connector. 
Nevertheless, the proposal in Zhou et al.~\cite{zhou2020practical} mainly focused on manipulating one pair of connectors instead of multi-connector manipulation, which is more common in actual production.

\subsection{Deep Learning-Based Object Detection}

The rebirth of convolutional neural networks (CNNs) in 2012~\cite{krizhevsky2012imagenet} initiated the research on introducing deep learning~\cite{lecun2015deep} to object detection~\cite{zou2023object}, which further promoted the remarkable development of two major groups of detectors for object detection based on deep learning in previous years: two-stage detection and one-stage detection~\cite{zou2023object}.

Similar to the attentional mechanism of the human brain, the two-stage detection model first scans the whole scenario coarsely and then focuses on regions of interest (ROIs) to distinguish the object~\cite{zhao2019object}.
The region-based convolutional neural network (R-CNN) proposed by Girshick et al.~\cite{girshick2014rich,girshick2015region} symbolized the inauguration of two-stage object detection.
In R-CNN~\cite{girshick2014rich,girshick2015region}, a set of object proposals were extracted and fed into a CNN model to extract features for classification.
However, the redundant feature computations due to many overlapped proposals made the detection speed extremely slow, which was improved later by Spatial Pyramid Pooling Networks (SPPNet)~\cite{he2015spatial}.
A Spatial Pyramid Pooling (SPP) layer was introduced in SPPNET~\cite{he2015spatial} to enable a CNN to generate a fixed-length representation to avoid re-scaling. Nevertheless, SPPNET~\cite{he2015spatial} remained multi-stage training and only fine-tuning fully-connected layers.
To improve R-CNN~\cite{girshick2014rich} and SPPNet~\cite{he2015spatial}, Fast R-CNN~\cite{girshick2015fast} was proposed later, where the detector and the bounding box regressor could be trained under the same network configurations simultaneously.
Furthermore, Faster R-CNN~\cite{ren2015faster} was proposed to accelerate the detection further by introducing a Region Proposal Network (RPN), but the problem of computation redundancy remained at the subsequent detection stage.
Besides the R-CNN family, Lin et al.~\cite{lin2017feature} proposed Feature Pyramid Networks (FPNs), which can be integrated into other detectors to enable high-level semantics building at all scales besides the feature maps of the networks' top layer.

Though able to attain high-precision detection, two-stage detection methods are constrained by their ponderous detection speed and computation, stimulating the research on one-stage detection.
You Only Look Once (YOLO)~\cite{redmon2016you} was the first deep learning-based one-stage detection that simultaneously predicted bounding boxes and probabilities for each sub-region of an image.
Although the detection speed was improved significantly, the localization accuracy dropped remarkably compared to two-stage detection, especially for some small objects, which was enhanced in YOLO's subsequent versions~\cite{redmon2018yolov3,bochkovskiy2020yolov4,redmon2017yolo9000,wang2022yolov7}.
There were also other one-stage detection methods besides the YOLO family proposed to improve the detection accuracy while maintaining the advantage of high detection speed, including Single-Shot Multibox Detector (SSD)~\cite{liu2016ssd}, RetinaNet~\cite{lin2017focal}, and CornerNet~\cite{law2018cornernet}.

Recent years have also witnessed the profound influence of Transformer models~\cite{vaswani2017attention} in deep learning and computer vision~\cite{khan2022transformers}, which has spawned DEtection TRansformer (DETR)~\cite{carion2020end} and Deformable DETR~\cite{zhu2021deformable} and promoted deep learning-based object detection to higher performance.

\section{THE DATASET OF CONNECTORS}
\label{sec:dataset}

\begin{figure}[t]
    \centering
    \includegraphics[width=\linewidth]{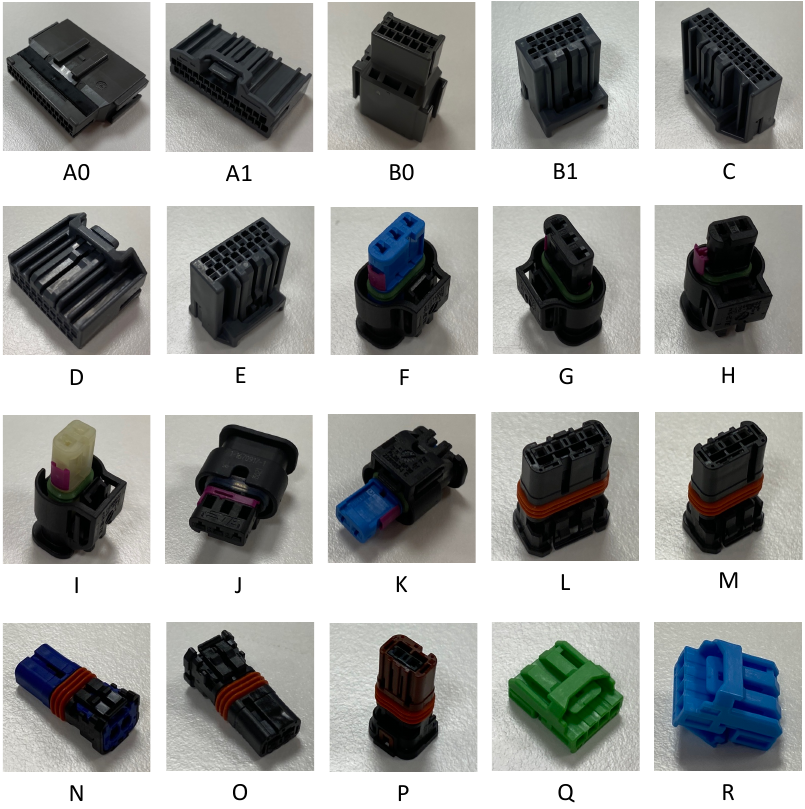}
    \caption{The twenty types of connectors collected for dataset creation. The class of each connector is simplified and labeled below images.}
    \label{fig:connector_20}
\end{figure}

The dataset is essential for learning-based object detection~\cite{everingham2009pascal,lin2014microsoft,russakovsky2015imagenet} and scalable deep learning-based solutions in industry~\cite{nguyen2022enabling}.
However, to the best of the authors' knowledge, there is no publicly available benchmark dataset dedicated to the detection of automotive wire harness connectors.
Thus, to facilitate the study of deep learning-based connector detection for the robotized assembly of automotive wire harnesses, a dataset was collected and annotated first, consisting of $20$ types of connectors commonly occurring on automotive wire harnesses installed in passenger vehicles.
Fig.~\ref{fig:connector_20} demonstrates one example image for each of the $20$ connectors.
The following subsections will introduce the strategy for image collection and annotation and summarize the statistics of the dataset used in the experiments.

\subsection{Image Collection Procedure}

Connectors are placed on a white workbench for image acquisition using the main camera of an iPhone 11.
The original image format is RGB, and each image has a size of $4032 \times 3024$ pixels.
The distance between the camera and connectors was not fixed, considering the various locations of connectors in the three-dimensional (3D) space in actual assembly situations.

\begin{figure}[t]
    \centering
    \includegraphics[width=\linewidth]{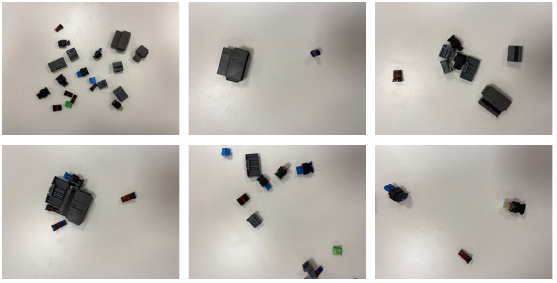}
    \caption{Examples of images with different combinations of connectors.}
    \label{fig:connector_mixed}
\end{figure}

There are $360$ images captured in total.
Initially, $60$ images of various combinations of connectors with random poses were collected to simulate the random distribution of connectors in the actual assembly scenario.
Fig.~\ref{fig:connector_mixed} demonstrates some examples of these $60$ images.
For clarification, the distribution of connectors in each of these $60$ images does not represent the actual distribution of connectors on practical automotive wire harnesses or in the final assembly of automotive wire harnesses.

\begin{figure}[t]
    \centering
    \includegraphics[width=\linewidth]{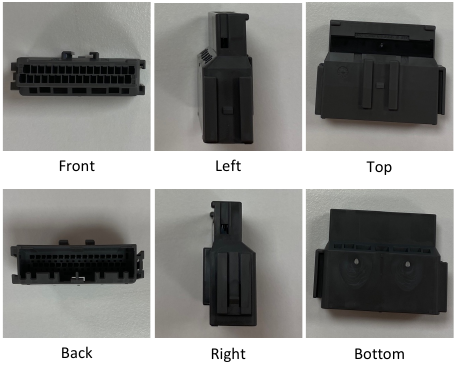}
    \caption{The six images captured from the front, back, top, down, left, and right of A0. These images are cropped from the raw data for demonstration.}
    \label{fig:A0_6}
\end{figure}

\begin{figure}[t]
    \centering
    \includegraphics[width=\linewidth]{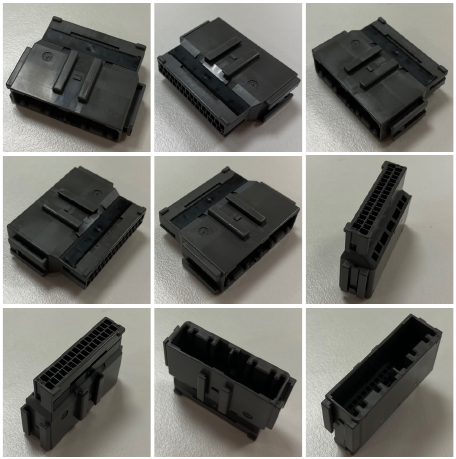}
    \caption{The other nine images of A0 were captured from random perspectives. These images are cropped from the raw data for demonstration.}
    \label{fig:A0_others}
\end{figure}

In addition, images of each of the $20$ connectors were also collected to train the detector with more features of respective classes.
For each connector, $15$ images were captured from different views, including six images captured from the front, back, top, down, left, and right of the connector, as an example of class A0 shown in Fig.~\ref{fig:A0_6}, and nine images captured from random perspectives, as an example of class A0 shown in Fig.~\ref{fig:A0_others}.

\subsection{Image Annotation Procedure}

The image annotation procedure of the dataset of connectors followed the methodology implemented in the PASCAL visual object classes (VOC) challenge 2007~\cite{everingham2009pascal}.

The image annotation includes the \textbf{class} and the \textbf{bounding box} for every connector in the target set of classes.
As shown in Fig.~\ref{fig:connector_20}, this study simplified the $20$ classes of connectors into A0, A1, B0, B1, C, D, E, F, G, H, I, J, K, L, M, N, O, P, Q, and R, which can be easily mapped to the actual types of connectors in practical applications.
An axis-aligned rectangular bounding box surrounding the connector was drawn for each connector visible in each image in the dataset.
Though relatively quick to annotate, choosing an axis-aligned rectangular bounding box for the annotation is a compromise.
Some connectors in images fit well because of their rectangular or approximately rectangular profiles, for example, class A0 shown in Fig.~\ref{fig:A0_6}.
However, for other connectors presented in images, an axis-aligned bounding box can be a poor fit because either they are not axis-aligned, for example, been captured from random perspectives (Fig.~\ref{fig:A0_others}) or placed randomly (Fig.~\ref{fig:connector_mixed}), or the connector is not in the shape of a box, for example, class I shown in Fig.~\ref{fig:connector_20}.

The actual image annotation was conducted using an annotation platform, Labelme~\cite{labelme}.
It was trivial to annotate images with a single connector due to the structured storage of images.
For images with multiple connectors, a list of visible connectors in each image was documented first during the image collection procedure.
Then, each connector visible in the images was compared with the original physical counterpart and annotated exhaustively.
The annotation results were compared to the documented list to guarantee the consistency and accuracy of the image annotation.

\subsection{Dataset Statistics}

\begin{figure}[t] 
    \centering
    \includegraphics[width=\linewidth]{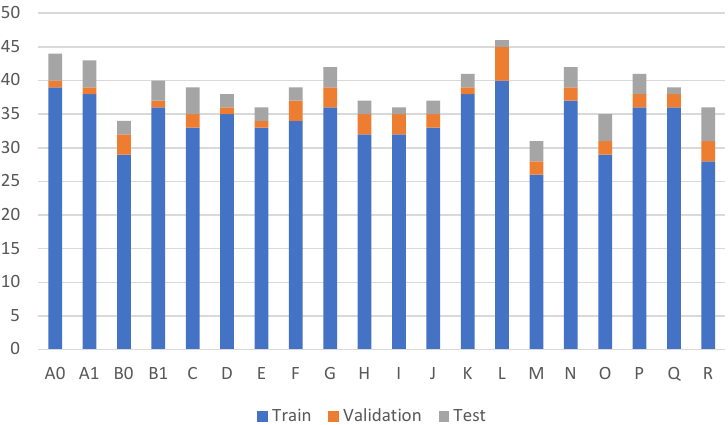}
    \caption{Histogram of the numbers of object instances shown in the collected connector dataset. The classes and the corresponding counts are shown on the x-axis and the y-axis, respectively.}
    \label{fig:dataset_hist}
\end{figure}

The total number of annotated images is $360$.
The data is primarily divided into three main subsets: training data (Train), validation data (Validation), and test data (Test), with a ratio of $90\%/5\%/5\%$.
The images in the validation set and test set were selected randomly.
For each subset of the connector dataset and class of connectors, the number of object instances is shown in TABLE~\ref{tab:stat}.
In the collected dataset, the most frequent class is ``L'', with $46$ object instances, and the least frequent class is ``M'', with $31$ object instances.
Fig.~\ref{fig:dataset_hist} illustrates a histogram of the number of object instances presented in different subsets of the collected connector dataset for each class of connectors.

\begin{table*}[htbp]
    \caption{The Numbers of Annotated Object Instances in the Collected Connector Dataset.}
    \label{tab:stat}
    \begin{center}
        \begin{tabular}{|c||c||c||c||c||c||c||c||c||c||c||c||c||c||c||c||c||c||c||c||c|}
            \hline
             & A0 & A1 & B0 & B1 & C & D & E & F & G & H & I & J & K & L & M & N & O & P & Q & R \\
            \hline
            Train & $39$ & $38$ & $29$ & $36$ & $33$ & $35$ & $33$ & $34$ & $36$ & $32$ & $32$ & $33$ & $38$ & $40$ & $26$ & $37$ & $29$ & $36$ & $36$ & $28$ \\
            Validation & $1$ & $1$ & $3$ & $1$ & $2$ & $1$ & $1$ & $3$ & $3$ & $3$ & $3$ & $2$ & $1$ & $5$ & $2$ & $2$ & $2$ & $2$ & $2$ & $3$ \\
            Test & $4$ & $4$ & $2$ & $3$ & $4$ & $2$ & $2$ & $2$ & $3$ & $2$ & $1$ & $2$ & $2$ & $1$ & $3$ & $3$ & $4$ & $3$ & $1$ & $5$ \\
            \hline
            Total & $44$ & $43$ & $34$ & $40$ & $39$ & $38$ & $36$ & $39$ & $42$ & $37$ & $36$ & $37$ & $41$ & $46$ & $31$ & $42$ & $35$ & $41$ & $39$ & $36$ \\
            \hline
        \end{tabular}
    \end{center}
\end{table*}

\begin{figure*}[htbp] 
    \centering
    \includegraphics[width=\linewidth]{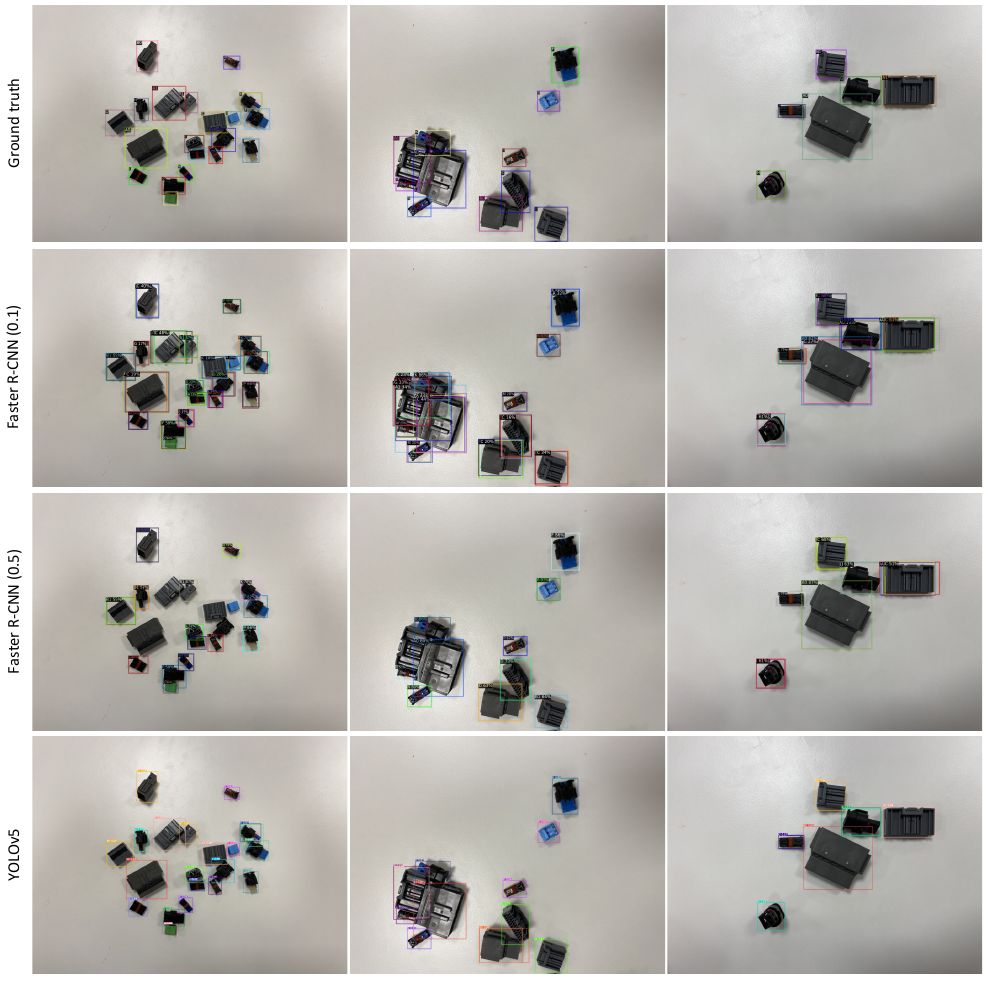}
    \caption{Detection results of Faster R-CNN~\cite{ren2015faster} with different threshold values and YOLOv5~\cite{yolov5} with bounding boxes around detected connectors and inferred classes on the upper left corners of bounding boxes. The first row demonstrates the corresponding ground truth. The second row demonstrates the Faster R-CNN~\cite{ren2015faster} detection results with a threshold of $0.1$. The third row demonstrates the Faster R-CNN~\cite{ren2015faster} detection results with a threshold of $0.5$. The last row demonstrates the YOLOv5~\cite{yolov5} detection results.}
    \label{fig:inference}
\end{figure*}

\section{EXPERIMENT SETTINGS}
\label{sec:exp}

This study investigated a two-stage detector and a one-stage detector for automotive wire harness connector detection.
The experiment on two-stage detection was conducted based on Faster R-CNN~\cite{ren2015faster}, and the one-stage detection was achieved based on YOLO~\cite{redmon2016you}.
Both models were trained using the union of the train and validation set of the collected connector dataset and evaluated on the test set using an NVIDIA GeForce RTX 4090.
The following subsections introduce the detailed implementation of the two-stage detection and the one-stage detection, respectively.

\subsection{Two-Stage Detection}

This study investigated the two-stage detection based on Faster R-CNN~\cite{ren2015faster} and implemented Faster R-CNN~\cite{ren2015faster} with ResNet~\cite{he2016deep} plus Feature Pyramid Network (FPN)~\cite{lin2017feature} as the backbone.
The overall baselines and hyper-parameters followed Faster R-CNN~\cite{ren2015faster} provided in the publicly available code of Detectron2~\cite{wu2019detectron2}.
Specifically, the model was trained with a learning rate of $0.00025$ using Stochastic Gradient Descent (SGD) as the optimizer.
The batch size was $8$.
The weights of the model were initiated with the pre-trained checkpoint, \textit{faster\_R-CNN\_R\_101\_FPN\_3x}, provided by Detectron2~\cite{wu2019detectron2}.

\subsection{One-Stage Detection}

YOLO~\cite{redmon2016you} was selected as the backbone of the one-stage detection in the experiment.
The overall baselines and hyper-parameters of the one-stage detection in this study followed the publicly available code of YOLOv5~\cite{yolov5}.
Specifically, the model was trained with an initial learning rate of $0.01$ using SGD as the optimizer.
The weight decay was $0.0005$, and the momentum was $0.937$.
The batch size was $16$.
The weights of the model were initiated with the pre-trained checkpoint, \textit{yolov5x}, provided by YOLOv5~\cite{yolov5}.
An early-stop module was adopted to control the end of the training process, which terminated the training if there was no improvement after $300$ consecutive epochs.

\section{RESULTS AND DISCUSSION}
\label{sec_result}

The initialization, training, and evaluation of the two-stage detection model based on Faster R-CNN~\cite{ren2015faster} and the one-stage detection model based on YOLOv5~\cite{yolov5} were conducted following the experiment protocol explained in section~\ref{sec:exp}.
Fig.~\ref{fig:inference} demonstrates some inference results of Faster R-CNN~\cite{ren2015faster} with two threshold values and YOLOv5~\cite{yolov5} as well as the corresponding ground-truth images with original bounding boxes and labels.

There are several sub-stages of processing involved in the Faster R-CNN~\cite{ren2015faster} algorithm.
In one of these sub-stages, which classifies regions of an image as either object or background, a threshold value is required to be set by the user to determine the confidence score needed for a region to be considered as an object, i.e., a region is considered as background and discarded if its confidence score is below the threshold value, otherwise, an object and retained.
In this study, two threshold values, $0.1$ and $0.5$, were set to evaluate the Faster R-CNN-based model, whose inference results are shown in the second and the third row of Fig.~\ref{fig:inference}.

As shown in the second row of Fig.~\ref{fig:inference}, all connectors are located with the threshold value of $0.1$, but there are many detection errors in the classes of connectors and uncertain detection of the positions of connectors in the detection results.
With the elevation of the threshold values to $0.5$, the inference results present a more precise detection on different connectors, as shown in the third row of Fig.~\ref{fig:inference}, but some bounding boxes are excluded from the final detection results, which left some connectors in images not detected, leading to a deteriorated recall rate.
In general, the detection results with both threshold values demonstrate the deep learning-based two-stage detection model's effectiveness on the task of detection on automotive wire harness connectors.
Nevertheless, more data on connectors is desired to train a better detection model with higher precision and recall rates.
Further study on selecting the appropriate threshold value is also critical to make the detection more accurate and robust for practical applications.

The last row in Fig.~\ref{fig:inference} demonstrates some detection results from the one-stage detection model based on YOLOv5~\cite{yolov5}.
The detection results indicate the effectiveness of the deep learning-based one-stage detection model on connector detection.
However, there are also some detection errors on the positions and classes of connectors presented in the detection results, where the augmentation of the dataset~\cite{everingham2015pascal} can be helpful for better training and inference.

\begin{table}[t]
    \caption{The Precision ($\%$) of Faster R-CNN~\cite{ren2015faster} with Threshold Values of $0.1$ and $0.5$ and YOLOv5~\cite{yolov5} Among Classes.}
    \label{tab:ap_classes}
    \begin{center}
        \begin{tabular}{|c||c||c||c||c||c|}
            \hline
             & A0 & A1 & B0 & B1 & C \\
            \hline
            Faster R-CNN~\cite{ren2015faster} ($0.1$) & $\textbf{85.1}$ & $69.9$ & $20.2$ & $68.0$ & $28.4$ \\
            Faster R-CNN~\cite{ren2015faster} ($0.5$) & $67.8$ & $43.1$ & $0.0$ & $59.7$ & $11.6$ \\
            YOLOv5~\cite{yolov5} & $76.4$ & $\textbf{79.2}$ & $\textbf{100.0}$ & $\textbf{73.8}$ & $\textbf{69.7}$ \\
            \hline
            \hline
             & D & E & F & G & H \\
            \hline
            Faster R-CNN~\cite{ren2015faster} ($0.1$) & $\textbf{12.6}$ & $16.4$ & $75.1$ & $\textbf{56.4}$ & $50.5$ \\
            Faster R-CNN~\cite{ren2015faster} ($0.5$) & $\textbf{12.6}$ & $0.0$ & $75.1$ & $\textbf{56.4}$ & $50.5$ \\
            YOLOv5~\cite{yolov5} & $0.0$ & $\textbf{47.2}$ & $\textbf{100.0}$ & $30.9$ & $\textbf{90.7}$ \\
            \hline
            \hline
             & I & J & K & L & M \\
            \hline
            Faster R-CNN~\cite{ren2015faster} ($0.1$) & $35.0$ & $\textbf{60.1}$ & $\textbf{80.1}$ & $45.0$ & $73.2$ \\
            Faster R-CNN~\cite{ren2015faster} ($0.5$) & $0.0$ & $35.3$ & $\textbf{80.1}$ & $45.0$ & $63.1$ \\
            YOLOv5~\cite{yolov5} & $\textbf{88.5}$ & $53.0$ & $63.2$ & $\textbf{85.5}$ & $\textbf{100.0}$ \\
            \hline
            \hline
             & N & O & P & Q & R \\
            \hline
            Faster R-CNN~\cite{ren2015faster} ($0.1$) & $91.1$ & $53.1$ & $76.6$ & $80.0$ & $82.9$ \\
            Faster R-CNN~\cite{ren2015faster} ($0.5$) & $91.1$ & $35.6$ & $56.4$ & $80.0$ & $69.7$ \\
            YOLOv5~\cite{yolov5} & $\textbf{91.6}$ & $\textbf{96.4}$ & $\textbf{94.6}$ & $\textbf{93.0}$ & $\textbf{96.8}$ \\
            \hline
        \end{tabular}
    \end{center}
\end{table}

\begin{table}[t]
    \caption{The Mean Average Precision ($\%$) of Faster R-CNN~\cite{ren2015faster} with Threshold Values of $0.1$ and $0.5$ and YOLOv5~\cite{yolov5}.}
    \label{tab:mAP}
    \begin{center}
        \begin{tabular}{|c||c||c|}
            \hline
             & mAP$_{50}$ & mAP$_{50-95}$ \\
            \hline
            Faster R-CNN~\cite{ren2015faster} ($0.1$) & $69.4$ & $58.0$ \\
            Faster R-CNN~\cite{ren2015faster} ($0.5$) & $54.4$ & $46.7$ \\
            YOLOv5~\cite{yolov5} & $\textbf{88.5}$ & $\textbf{82.1}$ \\
            \hline
        \end{tabular}
    \end{center}
\end{table}

\begin{figure*}[t]
    \centering
    \includegraphics[width=\linewidth]{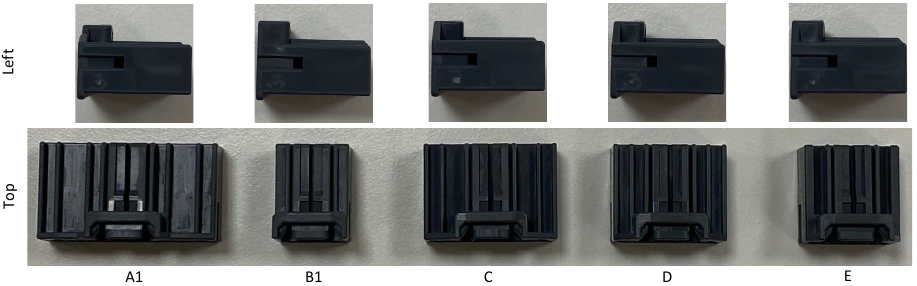}
    \caption{Class A1, B1, C, D, and E with highly similar profiles.}
    \label{fig:same_side}
\end{figure*}

\begin{figure*}[t]
    \centering
    \includegraphics[width=\linewidth]{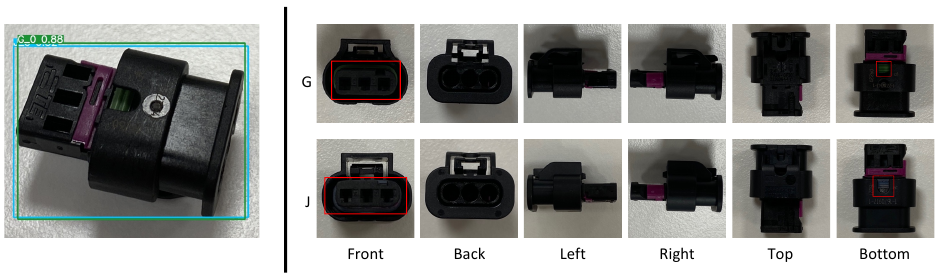}
    \caption{Inference result by YOLOv5 (left) on class G and J, whose exteriors are highly similar but the colors of seal rings inside are different (highlighted by red rectangles).}
    \label{fig:J_G}
\end{figure*}

Quantitatively, TABLE~\ref{tab:ap_classes} and TABLE~\ref{tab:mAP} present the rate of precision and mean Average Precision (mAP) of the Faster R-CNN-based model with threshold values of $0.1$ and $0.5$ and the YOLOv5-based model.
In general, the YOLOv5-based model outperforms the Faster R-CNN-based model regarding mAP on the collected connector dataset under the experiment settings in this study.
However, there are several rates of precision in TABLE~\ref{tab:ap_classes} lower than $50\%$, including the ones of both detection model on class D and E, the ones of the Faster R-CNN-based model on class C, and the one of the YOLOv5-based model on the class G.

By observing the exteriors of the connectors in the collected dataset, we find that similar designs of some connectors may affect detection performance.
For example, the widths of classes A1, B1, C, D, and E are different, but their left and right profiles are highly similar, as shown in Fig.~\ref{fig:same_side}, and classes G and J have identical exteriors but different seal rings inside the connectors, which are occluded when the images are captured from specific perspectives, as shown in Fig.~\ref{fig:J_G}.
These observations indicate that if some connectors share similar exterior designs and are placed with specific poses, their distinguishable features can be occluded, making it hard to recognize them.
Nonetheless, similar exteriors motivate two feasible strategies to relieve this detection problem: 1) conducting further connector detection based on multi-view images or videos; 2) re-design the exteriors of connectors with more distinguishable features.
Specifically, for the former solution, if the inference of the class of a connector is uncertain, multi-view images of the connector or a video capturing different views of the connector can be acquired for further classification.
And for the latter solution, changing the design of the exteriors of connectors, for example, changing the color of the whole connector or part of the connector, may substantially facilitate the detection, which calls for collaboration with the manufacturers of connectors.

\section{CONCLUSIONS AND FUTURE WORK}
\label{sec:conclu}

This study collects a dataset with twenty types of connectors commonly used on automotive wire harnesses and trains a two-stage Faster R-CNN-based detection model and a one-stage YOLOv5-based detection model to validate the feasibility of deep learning-based connector detection for robotized automotive wire harnesses assembly.
The experiment results indicate the effectiveness of both types of object detection methods and demonstrate the better performance achieved by the one-stage YOLOv5-based model on detecting automotive wire harness connectors but also reflect problematic detection outcomes that require further study with other detection algorithms and more data, which will be investigated in the future research.
In addition, observations on collected connectors motivate the problematic detection potentially affected by the similar designs of some connectors, especially the exteriors, which leads to future studies on multi-view image-based and video-based connector detection as well as on new exterior designs of connectors.







\bibliographystyle{IEEEtran}
\bibliography{IEEEabrv,mybib}

\end{document}